\documentclass[10pt,twocolumn,letterpaper]{article}

\usepackage[pagenumbers]{cvpr} 

\usepackage{graphicx}
\usepackage{amsmath}
\usepackage{amssymb}
\usepackage{booktabs}
\usepackage{xspace}
\usepackage{array}
\usepackage{makecell}
\usepackage{graphicx}
\usepackage{wrapfig}
\usepackage{caption}
\usepackage{xcolor}
\usepackage{algorithm}
\usepackage{listings}

\usepackage{pifont}
\newcommand{\xmark}{\ding{55}}%

\newcommand{\name}{SparseInst}

\newcommand{\tb}[1]{\textbf{#1}}

\renewcommand{\wrt}{\textit{w}.\textit{r}.\textit{t}. }

\newcommand{\apm}{AP}
\newcommand{\aps}{AP & AP$_{50}$ & AP$_{75}$}
\newcommand{\apl}{AP$_{50}$ & AP$_{75}$ & AP$_{S}$ & AP$_{M}$ & AP$_{L}$}
\newcommand{\tline}{\Xhline{1pt}}

\makeatletter\renewcommand\paragraph{\@startsection{paragraph}{4}{\z@}
  {.5em \@plus1ex \@minus.2ex}{-.5em}{\normalfont\normalsize\bfseries}}\makeatother

\newcommand{\tablestyle}[2]{\setlength{\tabcolsep}{#1}\renewcommand{\arraystretch}{#2}\centering}
\definecolor{citecolor}{HTML}{248B21}
\usepackage[pagebackref,breaklinks,citecolor=citecolor,colorlinks]{hyperref}

\usepackage[capitalize]{cleveref}
\crefname{section}{Sec.}{Secs.}
\Crefname{section}{Section}{Sections}
\Crefname{table}{Table}{Tables}
\crefname{table}{Tab.}{Tabs.}

\def\cvprPaperID{645} 
\def\confName{CVPR}
\def\confYear{2022}

\begin{document}

\title{Sparse Instance Activation for Real-Time Instance Segmentation}


\author{
{Tianheng Cheng} \textsuperscript{1, 2} \quad {Xinggang Wang} \textsuperscript{1}$^\dag$ \quad {Shaoyu Chen} \textsuperscript{1, 2} \quad {Wenqiang Zhang} \textsuperscript{1} \\
{Qian Zhang} \textsuperscript{2} \quad {Chang Huang} \textsuperscript{2} \quad {Zhaoxiang Zhang} \textsuperscript{3} \quad {Wenyu Liu} \textsuperscript{1}
\\
\textsuperscript{1} School of EIC, Huazhong University of Science \& Technology \quad \textsuperscript{2} Horizon Robotics \\
\textsuperscript{3} Institute of Automation, Chinese Academy of Sciences (CASIA) \\
{\tt\small \{thch,xgwang,shaoyuchen,wq\_zhang,liuwy\}@hust.edu.cn \quad \{qian01.zhang,chang.huang\}@horizon.ai} \\ 
{\tt\small zhaoxiang.zhang@ia.ac.cn}}

\maketitle

\begin{abstract}
In this paper, we propose a conceptually novel, efficient, and fully convolutional framework for real-time instance segmentation. Previously, most instance segmentation methods heavily rely on object detection and perform mask prediction based on bounding boxes or dense centers. In contrast, we propose a sparse set of instance activation maps, as a new object representation, to highlight informative regions for each foreground object. Then instance-level features are obtained by aggregating features according to the highlighted regions for recognition and segmentation. Moreover, based on bipartite matching, the instance activation maps can predict objects in a one-to-one style, thus avoiding non-maximum suppression (NMS) in post-processing. Owing to the simple yet effective designs with instance activation maps, SparseInst has extremely fast inference speed and achieves 40 FPS and 37.9 AP on the COCO benchmark, which significantly outperforms the counterparts in terms of speed and accuracy. Code and models
are available at \url{https://github.com/hustvl/SparseInst}.

\end{abstract}

\section{Introduction}
\let\thefootnote\relax\footnotetext{$^\dag$Xinggang Wang is the corresponding author.}

Instance segmentation aims to generate instance-level segmentation for each object in an image. Based on the advances in deep convolutional neural networks and object detection, recent works \cite{MaskRCNNHeGDG17,MSRCNNHuangHGHW19,HTCChenPWXLSF0SOLL19,BMaskChengWH020,SOLOWangKSJL20} have made tremendous progress in instance segmentation and achieved impressive results on large-scale benchmarks, \eg, COCO~\cite{COCOLinMBHPRDZ14}. However, developing real-time and efficient instance segmentation algorithms is still challenging and urgent, especially for autonomous driving and robotics.

\begin{figure}
\centering
\includegraphics[width=1.0\linewidth]{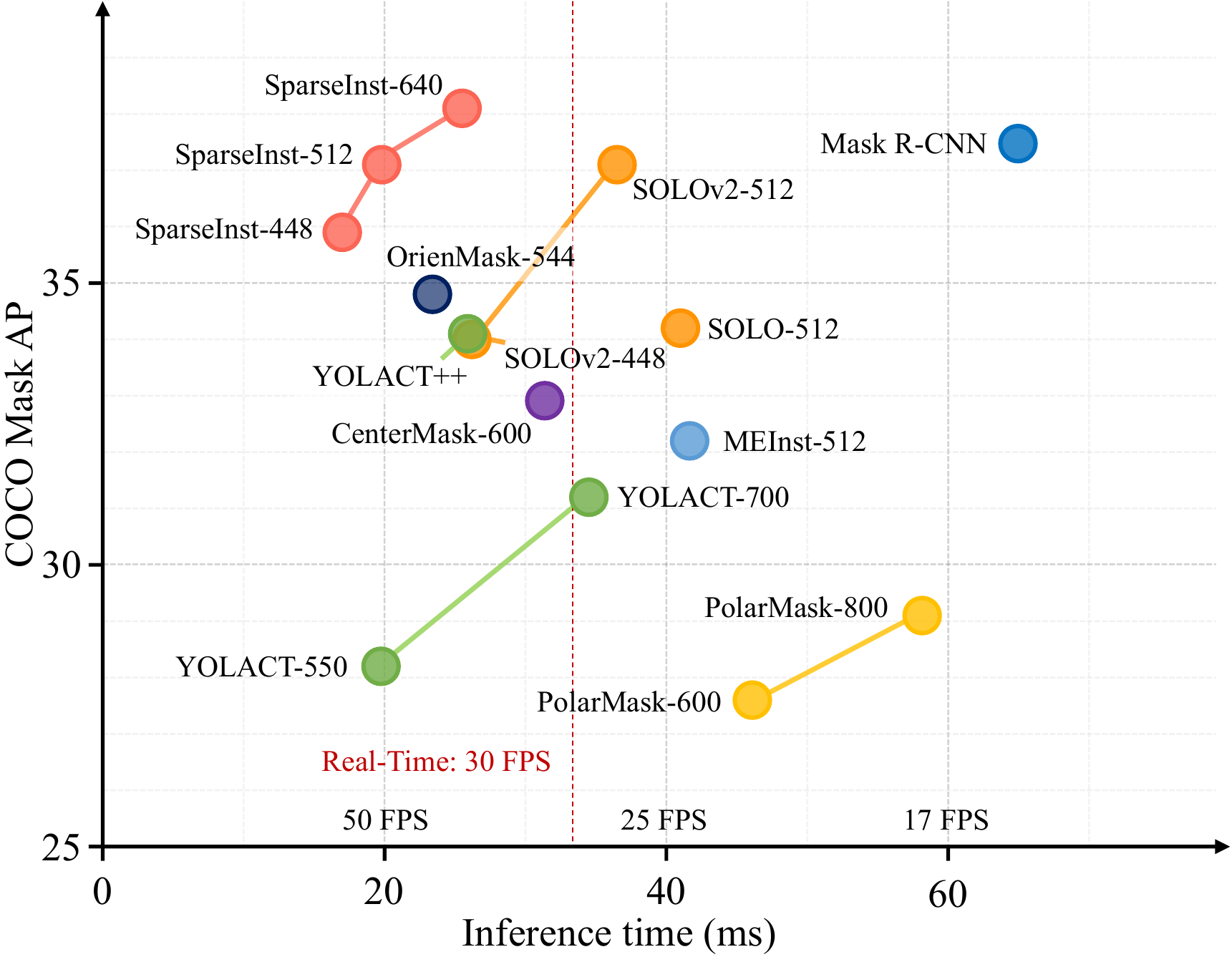}
\caption{\textbf{Speed-and-accuracy Trade-off.} The proposed SparseInst outperforms most state-of-the-art methods in both speed and accuracy for real-time instance segmentation. Inference speeds are measured on one NVIDIA 2080Ti.}
\label{fig:ap_latency}
\vspace{-16pt}
\end{figure}

Prevalent methods tend to adopt detectors~\cite{FRCNNRenHG017,FCOSTianSCH19} to localize instances first and then segment through region-based convolutional networks~\cite{MaskRCNNHeGDG17}, dynamic convolutions~\cite{TianSC20}, \etc.
Those methods are conceptually intuitive and achieve great performance.
However, when it comes to real-time instance segmentation, those methods suffer from some limitations.
Firstly, most methods employ dense anchors (centers) to localize and then segment objects, \eg, more than 5456 instances (given $512\times512$ input) in CondInst~\cite{TianSC20}, which incur lots of redundant predictions and much computation burden.
Besides, the receptive field of each pixel is limited and the contextual information is insufficient if we densely localize objects by centers or anchors~\cite{yolofabs-2103-09460,DensePointsYangXXZUWLH20}.
Secondly, most methods require multi-level prediction to handle the scale variation of natural objects, which inevitably increases the latency.
Region-based methods~\cite{MaskRCNNHeGDG17} apply RoI-Align to acquire region features, making it difficult to deploy algorithms to edge/embedded devices.
Finally, the post-processing also requires attention since the sorting and NMS as well as processing masks are time-consuming, especially for dense predictions. It's worth noting that even improved NMS~\cite{YolactBolyaZXL19,SOLOV2WangZKLS20} still takes $\sim$ 2ms, 10\% of total time.

\begin{figure}
    \centering
    \includegraphics[width=\linewidth]{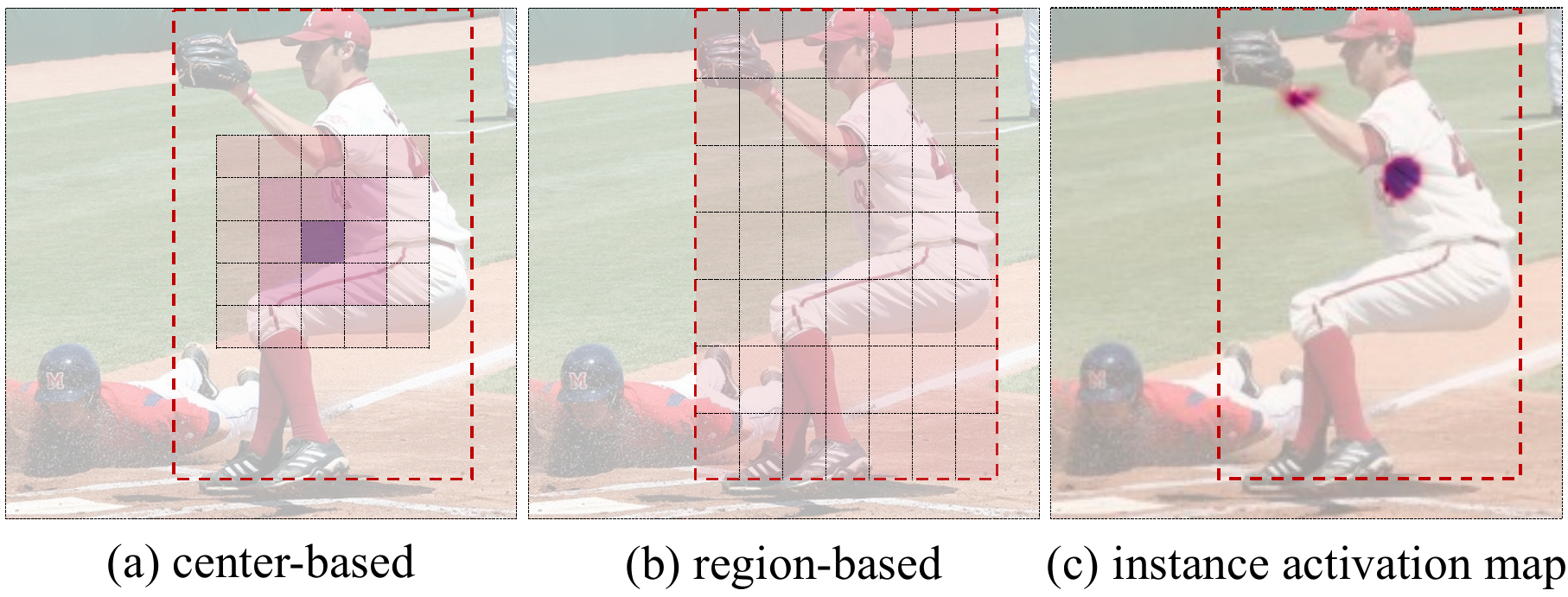}
    \caption{\textbf{Object Representation.} (a) center-based representation may fail to hit the instance;
    (b) region-based representation may contain features from other instances and background; 
    (c) instance activation map highlights instance-aware pixels.}
    \label{fig:represent_object}
    \vspace{-15pt}
\end{figure}

In this paper, we present a new \textit{highlight to segment} paradigm for real-time instance segmentation. 
Instead of using boxes or centers to represent objects, we exploit a sparse set of \textit{instance activation maps} (IAM) to highlight informative object regions, which is motivated by CAM~\cite{zhou2015cnnlocalization} widely used in weakly-supervised object localization.
Instance activation maps are instance-aware weighted maps and instance-level features can be directly aggregated according to the highlighted regions.
Then, recognition and segmentation are performed based on the instance features.
Figure~\ref{fig:represent_object} compares region-based, center-based, and IAM-based representations.
In comparison, IAM has the following advantages: (1) it highlights discriminative instance pixels, suppresses obstructive pixels, and conceptually avoids the incorrect instance feature localization problems in center-/region-based methods; (2) it aggregates instance features from the whole image and offers more contexts; (3) computing instance features with activation maps is rather simple without extra operation like RoI-Align~\cite{MaskRCNNHeGDG17}.
However, different from previous works~\cite{MaskRCNNHeGDG17,FCOSTianSCH19,SOLOV2WangZKLS20} using spatial priors (\ie, anchors and centers) to assign targets, instance activation maps are conditioned on the input and arbitrary for different objects and it is infeasible to assign targets with hand-crafted rules for training.
To address that, we formulate the label assignment for instance activation maps as a bipartite matching problem, which is recently proposed in DETR~\cite{DETRCarionMSUKZ20}.
Specifically, each target will be assigned to an object prediction as well as its activation map through Hungarian algorithm~\cite{HungarianStewartAN16}.
During training, the bipartite matching facilitates the instance activation maps to highlight individual objects and inhibit the redundant predictions, thus avoiding NMS during inference.

Further, we materialize this paradigm and propose \name, an extremely simple but efficient method for instance segmentation.
\name{} adopts single-level prediction and consists of a backbone to extract image features, an encoder to enhance the multi-scale representation for single-level features, and a decoder to compute the instance activation maps, perform recognition and segmentation, as shown in Figure~\ref{fig:main_arch}.
\name{} is a pure and fully convolutional framework and independent from detectors.
Benefiting from the facts: (1) the sparse predictions through the instance activation maps; (2) single-level prediction; (3) compact structures; (4) simple post-processing without NMS or sorting, \name{} has extremely fast inference speed and achieves 37.9 mask AP on MS-COCO \texttt{test-dev} with 40.0 FPS on one NVIDIA 2080Ti GPU,  outperforming most state-of-the-art methods for real-time instance segmentation. 
Given 448$\times$ input, \name{} achieves 58.5 FPS with competitive accuracy, which is faster than previous methods.
We hope the proposed \name~can serve as a general framework for (real-time) end-to-end instance segmentation.

\section{Related Work}
According to object representations, existing methods for instance segmentation can be divided into two groups, \ie region-based methods and center-based methods.
\paragraph{Region-based Methods.}
Region-based methods rely on object detectors, \eg, Faster R-CNN~\cite{FRCNNRenHG017}, to detect objects and acquire bounding boxes, and then apply RoI-Pooling~\cite{FRCNNRenHG017} or RoI-Align~\cite{MaskRCNNHeGDG17} to extract region features for pixel-wise segmentation. 
Mask R-CNN~\cite{MaskRCNNHeGDG17}, as the representative method, extends Faster R-CNN by adding a mask branch to predict masks for objects and offers a strong baseline for end-to-end instance segmentation.
\cite{PointRendKirillovWHG20,BMaskChengWH020,SegFixYuanXCW20,boundarypatchTangCLLZH21} address the low-quality segmentation and coarse boundaries arising in Mask R-CNN and present several approaches to refine the mask predictions for high-quality masks.
\cite{CascdeCaiV21,ChenPWXLSF0SOLL19} exploit cascade structures to progressively improve the object localization for more accurate mask prediction.
\paragraph{Center-based Methods.}
Recently, many approaches employ the single-stage detectors, especially the anchor-free detectors~\cite{FCOSTianSCH19}. 
These approaches represent objects by center pixels instead of bounding boxes and segment using the center features.
Several methods~\cite{PolarMaskXieSSWLLSL20,ESEXuWQL19} explores the object contours but show some limitations for objects having hollows or multiple parts.
YOLACT~\cite{YolactBolyaZXL19} generates instance masks by the assembly of mask coefficients and prototype masks.
MEInst~\cite{MEInstZhangTSYY20} and CondInst~\cite{TianSC20} extend FCOS~\cite{FCOSTianSCH19} by predicting the encoded mask vector or mask kernels for dynamic convolution~\cite{DynamicConvChenDLCYL20} respectively.
SOLO~\cite{SOLOWangKSJL20,SOLOV2WangZKLS20}, as a detector-free method, yet localize and recognize objects by centers as well as generating the mask kernels.
The proposed \name{} exploits sparse instance activation maps to represent objects with a simple pipeline and high efficiency.

\paragraph{Bipartite Matching for Object Detection.}
The bipartite matching has been widely explored for end-to-end object detection~\cite{DETRCarionMSUKZ20,DeDETRZhuSLLWD21,HungarianStewartAN16,sparsercnnabs-2011-12450,peize2020onenet,abs-2012-03544,OneNetSunJXSYWL21}, which avoids NMS in post-processing. 
Recently, SOLQ~\cite{SOLQabs-2106-02351} and ISTR~\cite{ISTRabs-2105-00637} exploit the mask encodings for instance segmentation.
QueryInst~\cite{queryinstabs-2105-01928} extends \cite{sparsercnnabs-2011-12450} by adding dynamic mask heads.
Besides, \cite{segformer_abs-2109-03814,mask_former_abs-2107-06278,WangZAYC21MaxDeepLab,KNetabs-2106-14855} employ transformers with instance and semantic queries to obtain panoptic segmentation results.
However, our method aiming at fast speed is motivated by the instance activation maps as object representation for instance-level recognition and segmentation. 
And the concise yet effective representation drives the framework rather fast.


\section{Method}

\begin{figure*}
    \centering
    \includegraphics[width=0.85\linewidth]{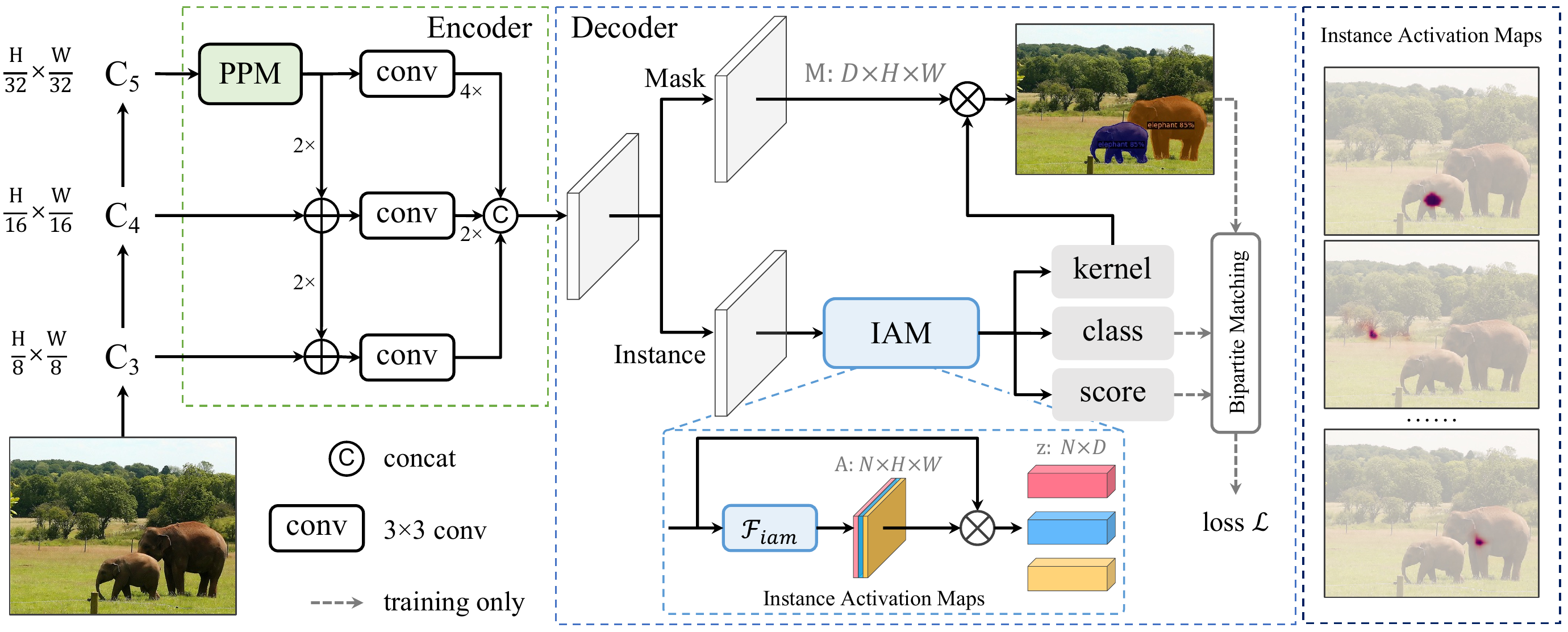}
    \vspace{-5pt}
    \caption{\textbf{The architecture of \name.} \name~contains three main components: \textit{backbone}, \textit{encoder} and \textit{IAM-based decoder}. 
    Given the input image, the backbone extracts the multi-scale image features (\ie, \{C$_3$,C$_4$,C$_5$\}). The encoder employs pyramid pooling module (PPM)~\cite{ZhaoSQWJ17} to enlarge the receptive field and fuses the multi-scale features. `$4\times$'or `$2\times$' denote the upsampling by a factor 4 or 2. The IAM-based decoder consists of two branches, \ie an instance branch and a mask branch. In the instance branch, the `IAM' module predicts the instance activation maps (shown in the right column) to acquire the instance features $\{z_i\}^N$ for recognition and mask kernels. The mask branch aims to provide mask features $\mathbf{M}$ and  will be multiplied with the predicted kernels to generate segmentation masks.}
    \label{fig:main_arch}
    \vspace{-10pt}
\end{figure*}

In this section, we first investigate the instance activation maps for representing objects.
Then we present a novel framework which exploits the sparse set of instance activation maps to highlight objects and aggregate instance features for instance-level recognition and segmentation.

\subsection{Instance Activation Maps}
\label{sec:instance_activaion_maps}
\paragraph{Formulation.}
Intuitively, instance activation maps are instance-aware weighted maps which aim to highlight the informative regions for each object.
And the features from the highlighted regions are semantically abundant and instance-aware for both recognizing and separating objects. 
Therefore, we directly aggregate the features according to the  activation maps as the instance features.
Given the input image features $\mathbf{X} \in \mathbb{R}^{D\times (H\times W)}$,  instance activation maps can be formulated as: $\mathbf{A} \!=\! \mathcal{F}_{iam}(\mathbf{X}) \in \mathbb{R}^{N\times (H\times W)}$,
where $\mathbf{A}$ is the sparse set of $N$ instance activation maps and  $\mathcal{F}_{iam}(\cdot)$ is a simple network with a sigmoid non-linearity.
Then we can obtain the sparse set of instance features by gathering distinctive information from the input feature maps $\mathbf{X}$ with the instance activation maps through: $z \!=\! \bar{\mathbf{A}}\cdot\mathbf{X}^T \in \mathbb{R}^{N\times D}$,
where $z=\{z_i\}^N$ are the feature representations for $N$ potential objects in the image and $\bar{\mathbf{A}}$ is normalized to 1 for each instance map. The sparse instance-aware features $\{z_i\}^N$ are straightforwardly used for consequent recognition and instance-level segmentation.

\paragraph{Learning Instance Activations.}
\label{learn_iam}
Instance activation maps don't exploit explicit supervisions, \eg, instance masks, for learning to highlight objects.
Essentially, the subsequent modules for recognition and segmentation provide instance activation maps with indirect supervisions, which encourage the $\mathcal{F}_{iam}$ to discover informative regions.
Additionally, the supervisions are instance-aware due to the bipartite matching, which further enforces the $\mathcal{F}_{iam}$ to discriminate objects and activate only one object per map.
Consequently, the proposed instance activation maps are capable to highlight discriminative regions for individual objects.


\subsection{\name}
As illustrated in Figure~\ref{fig:main_arch}, \name~is a simple, compact, and unified framework which consists of a backbone network, an instance context encoder, and an IAM-based decoder. The backbone network, \eg, ResNet~\cite{HeZRS16}, extracts multi-scale features from the given image. The instance context encoder is attached to the backbone to enhance more contextual information and fuse the multi-scale features.
For faster inference, the encoder outputs single-level features of $\frac{1}{8}\times$ resolution \wrt the input image, and the features will be fed to subsequent IAM-based decoder to generate instance activation maps to highlight foreground objects for classification and segmentation. 

\subsection{Instance Context Encoder}
Objects in natural scenes tend to have wide range of scales, which is prone to degrade the performance of detectors.
Most approaches adopt multi-scale feature fusions, \eg, feature pyramids~\cite{LinDGHHB17}, and multi-level prediction to facilitate the recognition for objects of different scales.
Nevertheless, using multi-level pyramidal features increase the computation burden, especially for detectors using heavy heads~\cite{FocalLinGGHD17,FCOSTianSCH19}, as well as producing amounts of duplicate predictions.
Conversely, our method aiming at faster inference leverages single-level prediction.
Considering the limitations of the single-level features for objects of various scales, we reconstruct the feature pyramid networks and present an instance context encoder, as illustrated in Figure~\ref{fig:main_arch}.
The instance context encoder adopts a pyramid pooling module~\cite{ZhaoSQWJ17} after C$_5$ to enlarge the receptive fields and fuses features from P$_3$ to P$_5$ to further enhance the multi-scale representations for the output single-level features.

\subsection{IAM-based Segmentation Decoder}
Figure~\ref{fig:main_arch} illustrates the IAM-based segmentation decoder which contains an instance branch and a mask branch. The two branches are composed of a stack of $3\times3$ convolutions with 256 channels. The instance branch aims to generate instance activation maps and N instance features for recognition and instance-aware kernel. The mask branch is designed to encode instance-aware mask features. 

\paragraph{Location-Sensitive Features.}
\label{sec:location_sensitive}
Empirically, objects are localized in different positions and the spatial locations can be used as cues to distinguish instances.
Hence, we construct two-channel coordinate features which consists of normalized absolute $(x,y)$ coordinates of spatial locations, which is similar to CoordConv~\cite{CoordConvLiuLMSFSY18}. Then we concatenate the output features from the encoder with coordinate features to enhance the instance-aware representation.

\paragraph{Instance Activation Maps $\mathcal{F}_{iam}$.}
We adopt a simple yet effective $3\!\times\!3$ convolution with sigmoid as the vanilla $\mathcal{F}_{iam}$, which highlights each instance with a single activation map.
Accordingly, instance features $\{z_i\}$ are obtained through activation maps, in which each potential object is encoded into a 256-d vector.
Then three linear layers are applied for classification, objectness score, and mask kernel $\{w_i\}^N$.
Further, to obtain fine-grained instance features, we present the \textit{group instance activation maps} (Group-IAM) to highlight a groups of regions for each object, \ie, multiple activation maps per object.
Specifically, we adopt a 4-group $3\!\times\!3$ convolution as the $\mathcal{F}_{iam}$ for Group-IAM and aggregate instance features by concatenating features from a group.

\paragraph{IoU-aware Objectness.}
\label{iou_objectness}
We discover that the one-to-one assignment will enforce most predictions to be background which may lower the classification confidence and cause misalignments between classification scores and segmentation masks.
To alleviate the above issues, we introduce the IoU-aware objectness to adjust the classification outputs.
We adopt the estimated IoU between predicted masks and ground-truth masks as the targets for foreground objects.
The ground-truth objectness for instances is varied and can facilitate the network to separate instances.
Different from ~\cite{MSRCNNHuangHGHW19}~using an extra head to predict IoU score based on mask predictions, we only adopt IoUs as the objectness targets.
At inference stage, we rescore the classification probability $p_i$ with the IoU-aware objectness $s_i$ and obtain the ultimate probability $\tilde{p_i} = \sqrt{p_i\cdot s_i}$, where $i$ denotes the $i$-th instance.

\paragraph{Mask Head.}
With the instance-aware mask kernels $\{w_i\}^N$ generated by the instance branch, the segmentation mask for each instance can be directly produced by $m_i = w_i\cdot \mathbf{M}$, where $m_i$ is the $i$-th predicted mask and its corresponding kernel is $w_i \in \mathbb{R}^{1\times D}$. $\mathbf{M} \in \mathbb{R}^{D\times H\times W}$ is the mask features. The final segmentation mask will be upsampled (via bilinear interpolation) to $1\times$ \wrt original resolution.


\subsection{Label Assignment and Bipartite Matching Loss}

The proposed \name{} outputs a fixed-size set of predictions and it's difficult to assign ground-truth objects with hand-crafted rules.
To tackle the end-to-end training, we formulate the label assignment as bipartite matching~\cite{DETRCarionMSUKZ20}.
Firstly, we propose a pairwise dice-based \textit{matching score} $\mathcal{C}(i, k)$ for $i$-th prediction and $k$-th ground-truth object in Eq.~\eqref{matching}, which is determined by classification scores and dice coefficients of segmentation masks.
\begin{equation}
    \label{matching}
    \mathcal{C}(i, k) = p_{i,c_k}^{1-\alpha}\cdot\text{DICE}(m_i,t_k)^{\alpha},
\end{equation}
where $\alpha$ is a hyper-parameter to balance the impacts of classification and segmentation and empirically set to 0.8. $c_k$ is termed as the category label for the $k$-th ground-truth object and $p_{i,c_k}$ indicates the probability for the category $c_k$ of $i$-th prediction. $m_i$ and $t_k$ are the masks of $i$-th prediction and $k$-th ground-truth object respectively.
The dice coefficient is defined in Eq.~\eqref{dice}.
\begin{equation}
    \label{dice}
    \text{DICE}(m,t) = \frac{2\sum_{x,y} m_{xy} \cdot t_{xy}}{\sum_{x,y}m^2_{xy}+\sum_{x,y}t^2_{xy}},
\end{equation}
where $m_{xy}$ and $t_{xy}$ denote the pixels at $(x,y)$ in the predicted mask $m$ and ground-truth mask $t$ respectively.
Then, we adopt Hungarian algorithm~\cite{HungarianStewartAN16} to find the optimal match between $K$ ground-truth objects and $N$ predictions. 
%

The training loss is defined in Eq.~\eqref{loss}, involving losses for classification, objectness prediction, and segmentation.
\begin{equation}
    \label{loss}
    \mathcal{L} = \lambda_c\cdot\mathcal{L}_{cls} + \mathcal{L}_{mask} + \lambda_{s}\cdot\mathcal{L}_{s},
\end{equation}
where $\mathcal{L}_{cls}$ is focal loss~\cite{FocalLinGGHD17} for object classification, $\mathcal{L}_{mask}$ is the mask loss and $\mathcal{L}_{s}$ is the binary cross entropy loss for the IoU-aware objectness.
Considering the severe imbalance problem between background and foreground in full-resolution instance segmentation, we adopt a hybrid mask loss in Eq.~\eqref{loss_mask} by combining the dice loss~\cite{MilletariNA16} and pixel-wise binary cross entropy loss for segmentation mask.
\begin{equation}
    \label{loss_mask}
    \mathcal{L}_{mask} = \lambda_{dice}\cdot\mathcal{L}_{dice} +\lambda_{pix}\cdot\mathcal{L}_{pix},
\end{equation}
where $\mathcal{L}_{dice}$ and $\mathcal{L}_{pix}$ are dice loss and binary cross entropy loss,  $\lambda_{dice}$ and $\lambda_{pix}$ are corresponding coefficients.




\subsection{Inference}
The inference stage of \name~is much straightforward and concise. Forward the given images through the whole network and we can directly obtain $N$ instances with classification scores $\{\tilde{p_i}\}^N$ and corresponding raw segmentation masks $\{m_i\}^N$. Then we can determine the category and confidence score for each instance and obtain the final binary mask by thresholding. Sorting and NMS are not needed, thus making the inference procedure very fast.

\section{Experiments}
\label{expr}
In this section, we evaluate the accuracy and inference speed of our proposed \name~on the challenging MS-COCO dataset and provide detailed ablation studies about our framework as well as qualitative results.

\paragraph{Dataset and Evaluation Metrics.} Our experiments are conducted on the COCO dataset \cite{COCOLinMBHPRDZ14} which consists of 118k images for training, 5k for validation and 20k for testing.
All models are trained on \texttt{train2017} and evaluated on \texttt{val2017}. As for instance segmentation, we mainly report the \apm~for segmentation mask. For inference speed, we measure the frames per second (FPS) including the post-processing on one NVIDIA 2080Ti GPU.
TensorRT or FP16 is not used for acceleration.

\paragraph{Implementation Details.}
\label{expr_details}
\name~is built on  Detectron2~\cite{wu2019detectron2} and trained over 8 GPUs with a total of 64 images per mini-batch. Following the training schedule in \cite{peize2020onenet}, we adopt AdamW~\cite{LoshchilovH19} optimizer with a small initial learning rate $5\times10^{-5}$ with weight decay 0.0001. All models are trained for 270k iterations and learning rate is divided by 10 at 210k and 250k respectively. The backbone is initialized with the ImageNet-pretrained weights with frozen batchnorm layers and other modules are randomly initialized. We adopt random flip and scale jitter in training. The shorter side of images are randomly sampled from 416 to 640 pixels, while the longer side is less or equal to 864. Unless specified, we evaluated the speed and accuracy with the shorter size 640. Loss coefficients $\lambda_c$, $\lambda_{dice}$, $\lambda_{pix}$, and $\lambda_{s}$ are empirically set to 2.0, 2.0, 2.0, and 1.0 respectively. We adopt N=100 instances for each image. Besides, we provide a MindSpore \cite{mindspore} implementation of \name{}.

\subsection{Main Results}

\begin{table*}[htb]
    \centering
    \small
    \renewcommand{\arraystretch}{4pt}
    \renewcommand\arraystretch{1.1}
    \scalebox{0.90}{
    \begin{tabular}{l|l|cl|ccc|ccc}
    method & backbone & size & FPS  & \apm & \apl \\
    \tline
    
    MEInst~\cite{MEInstZhangTSYY20}   & R-50-FPN & 512  & 24.0  & 32.2 & 53.9 & 33.0 & 13.9    & 34.4 & 48.7 \\
    CenterMask~\cite{CenterMaskLeeP20} & R-50-FPN & 600 & 31.9 & 32.9 & - & - & 12.9 & 34.7 & 48.7  \\
    CondInst~\cite{TianSC20} & R-50-FPN & 800 & 20.4$^\dag$ & 35.4 & 56.4 & 37.6 & 18.4 & 37.9 & 46.9 \\
    SOLO~\cite{SOLOWangKSJL20}    & R-50-FPN & 512 & 24.4 & 34.2 & 55.9 & 36.0 & - & - & -  \\
    SOLOv2-Lite~\cite{SOLOWangKSJL20}    & R-50-FPN & 448 & 38.2 & 34.0 & 54.0 & 36.1 & 10.3 & 36.3 & 54.4  \\
    SOLOv2-Lite~\cite{SOLOWangKSJL20}    & R-50-DCN-FPN & 512 & 28.2 & 37.1 & 57.7 & 39.7 & 12.9 & \tb{40.0} & \tb{57.4}  \\
    PolarMask~\cite{PolarMaskXieSSWLLSL20} & R-50-FPN & 600 & 21.7$^\dag$ & 27.6 & 47.5 & 28.3 & 9.8 & 30.1 & 43.1 \\
    PolarMask~\cite{PolarMaskXieSSWLLSL20} & R-50-FPN & 800 & 17.2$^\dag$ & 29.1 & 49.5 & 29.7 & 12.6 & 31.8 & 42.3 \\
    YOLACT~\cite{YolactBolyaZXL19}    & R-50-FPN & 550  & 50.6  & 28.2 & 46.6 & 29.2 & 9.2 & 29.3 & 44.8 \\ 
    YOLACT~\cite{YolactBolyaZXL19}    & R-101-FPN & 700 & 29.0 & 31.2  & 50.6 & 32.8 & 12.1 & 33.3 & 47.1 \\
    YOLACT++~\cite{YolactBolyaZXL19} & R-50-DCN-FPN & 550 & 38.6 & 34.1 & 53.3 & 36.2 & 11.7 & 36.1 & 53.6 \\
    OrienMask~\cite{orien_maskabs-2106-12204} & D-53-FPN & 544 & 42.7 & 34.8 & 56.7 & 36.4 & \tb{16.0} & 38.2 & 47.8 \\ 
    \hline
    \textbf{\name} & R-50 & 608 & 44.6 & 34.7 & 55.3 & 36.6 & 14.3 & 36.2 & 50.7 \\
    \textbf{\name} & R-50-DCN & 608 & 41.6 & 36.8 & 57.6 & 38.9 & 15.0 & 38.2 & 55.2 \\
    \textbf{\name} & R-50-d & 608 & 42.8 & 36.1 & 57.0 & 38.2 & 15.0 & 37.7 & 53.1 \\
    \textbf{\name} & R-50-d-DCN & 608 & 40.0 & \tb{37.9} & \tb{59.2} & \tb{40.2} & 15.7 & 39.4 & 56.9 \\
    \end{tabular}}
    \vspace{-5pt}
    \caption{\textbf{COCO Instance Segmentation.} Comparisons with state-of-the-art methods for mask AP and speed on COCO \texttt{test-dev}. Inference speeds of all models are tested on our machine with one NVIDIA RTX 2080Ti except those marked with $^\dag$, which are inherited from their publications. }
    \label{tab:main_experiments}
    \vspace{-10pt}
\end{table*}


Since the SparseInst aims for real-time instance segmentation, we mainly compare \name~with the state-of-the-art methods towards real-time instance segmentation with respect to accuracy and inference speed. Results are evaluated on COCO \texttt{test-dev}. 
We provide \name~with group instance activation maps and different backbones to achieve the trade-off between speed and accuracy. 
We adopt ResNet-50~\cite{HeZRS16} to reach higher inference speed and its variant ResNet-d~\cite{he2018bag} to achieve better accuracy but with higher latency and aim for providing a stronger baseline for real-time instance segmentation.
Additionally, we adopt a simple random crop and larger weight decay (0.05) to better compare with OrienMask~\cite{orien_maskabs-2106-12204} and YOLACT~\cite{YolactBolyaZXL19}.
Table~\ref{tab:main_experiments} shows that our \name{} is superior to most real-time methods with better performance and faster inference speed.
SparseInst outperforms the popular real-time approach YOLACT by a remarkable margin with faster speed. 
Figure~\ref{fig:ap_latency} illustrates the speed-accuracy trade-off curve and the proposed SparseInst with R50-d and DCN~\cite{DCNZhuHLD19} obtains better trade-off compared with the counterparts and achieves 58.5 FPS and 35.5 mask AP with 448$\times$ input, which is superior to most real-time methods ($\ge$ 30FPS).



\subsection{Ablation Experiments}
We conduct a series of ablations to investigate \name, including experimental details about the components.

\paragraph{Instance Context Encoder.}
Table~\ref{tab:ablation_encoder} shows the impacts of the modifications to the vanilla feature pyramids~\cite{LinDGHHB17}.
Adding the pyramid pooling module for larger receptive fields and more object contexts brings significant improvement by 1.5 AP and 2.2 AP for larger objects (AP$_L$) while incurs negligible latency.
Moreover, fusing the multi-scale features from P$_3$ to P$_5$ further enhances the multi-scale feature representation and improves the performance by 0.7 AP and 2.0 AP$_L$.
The context encoder is rather essential for single-level prediction to cope with the limited receptive fields and provide better multi-scale features, thus bridging the gap between multi-level and single-level methods.
\begin{table}[]
    \centering
    \tablestyle{3.5pt}{1.1}
    \small
    \scalebox{0.9}{
    \begin{tabular}{cc|c|ccc|ccc}
    w/ fusion & w/ PPM & \textit{t}~(ms) & \aps  & AP$_{S}$ & AP$_{M}$ & AP$_{L}$ \\
    \tline
    & & \textbf{22.0} & 29.8 & 48.7 & 31.0 & 12.0 & 31.8 & 44.1 \\
     & \checkmark & 22.2 & 31.3 & 50.8 & 32.4 & \tb{14.0} & 33.2 & 46.2 \\
    \checkmark &  & 22.8 & 30.3 & 49.5 & 31.6 & 12.5 & 32.3 & 45.9 \\
    \checkmark & \checkmark & 22.9 & \tb{32.0} & \tb{52.0} & \tb{33.3} & 13.1 & \tb{34.5} & \tb{48.2} \\
    \end{tabular}}
    \caption{\textbf{Ablation on the Instance Context Encoder.} The vanilla encoder~\cite{LinDGHHB17} is incapable for single-level prediction. Leveraging PPM can enlarge the receptive fields and significantly improve the overall performance and adding multi-scale fusion further improves the accuracy, especially for AP$_L$. Notably, the extra latency of the improved encoder compared to the vanilla one is negligible.}
    \label{tab:ablation_encoder}
    \vspace{-5pt}
\end{table}

\paragraph{Structure of the Decoder.}
In Table~\ref{tab:ablation_decoder_structure}, we compare different structures of the two branches in the IAM-based Decoder.
We adopt 4 conv layers with 256 channels as the basic setting for both branches and evaluate the performance of models with different depths or widths.
Reducing width or reducing depth will lower the performance but increase the inference speed and it's worth noting that reducing channels to 128 performs worse.
Increasing the depth from 4 to 6 brings 0.4 AP improvement.
Considering the trade-off between speed and accuracy, we adopt width=256 and depth=4 in all experiments. 
Adding coordinate features improves the baseline by 0.5 AP with negligible time consumption, which indicates the effect of the explicit location-aware features as discussed in $\S$\ref{sec:location_sensitive}.
Table~\ref{tab:ablation_decoder_structure} also shows the effects of replacing the last convolution of the two branches with a deformable convolution. 
Using deformable convolution~\cite{DCNZhuHLD19} is optional and improves larger objects by enlarging the receptive field but consumes much time (+1.7ms).

\begin{table}
    \centering
    \small
    \renewcommand\tabcolsep{3.5pt}
    \renewcommand\arraystretch{1.1}
    \scalebox{0.90}{\begin{tabular}{cc|c|c|cccc|c}
    depth & width & coord? & dconv? & \apm & AP$_S$ & AP$_M$ & AP$_L$ & \textit{t} (ms)\\
    \tline
    4 & 256 & & & 31.5 & 13.4 & 33.5 & 47.9 & 22.9 \\
    4 & 256 & \checkmark & & 32.0 & 13.0 & 34.5 & 48.2 & 22.9\\
    4 & 256 & \checkmark & \checkmark & \tb{32.6} & 13.1 & 34.8 & \tb{49.2} & 24.6 \\
    \hline
    2 & 256 & \checkmark & & 31.0 & 12.9 & 33.2 & 47.0 & 20.6 \\
    6 & 256 & \checkmark & & 32.4 & \tb{13.7} & \tb{35.4} & 47.9 & 25.5 \\
    4 & 128 & \checkmark & & 30.6 & 12.4 & 32.5 & 46.2 & \tb{19.7} \\
    \end{tabular}}
    \captionof{table}{\textbf{Ablation on the structure of the decoder.} `coord.' denotes coordinates and `dconv.' denotes deformable convolution. Adding coordinates brings 0.5 AP improvement but with negligible latency. Replacing the last convolution with deformable convolution gives significant improvement on larger objects (AP$_L$). Reducing the width or depth improves the inference speed but lower the performance, while increasing the depth can further improve the accuracy but lower the speed.}
    \label{tab:ablation_decoder_structure}
    \vspace{-12pt}
\end{table}

\paragraph{Instance Activation Maps.}
$\mathcal{F}_{iam}$ is the key component for highlighting object regions, and we explore different designs for $\mathcal{F}_{iam}$ in Table~\ref{tab:ablation_iam}. Using softmax or $1\!\times\!1$ conv brings 0.4 AP and 1.2 AP drop, respectively.
Sigmoid (w/ norm) and softmax can be formulated as $s_i\!=\!\frac{f(x_i)}{\sum_kf(x_k)}$ where $f(x)\!=\!e^x$ for softmax and $f(x)\!=\!\frac{1}{1+e^{-x}}$ for sigmoid, which tends to saturate thus activate larger regions then softmax.
Adding extra $3\!\times\!3$ conv brings no gain but increases the computation cost.
Further, we evaluate the Group-IAM with different groups and Table~\ref{tab:ablation_iam} shows that using 4 groups improves the model by 0.7 AP.

\begin{table}
    \centering
    \renewcommand\arraystretch{1.1}
    \small
    \setlength{\tabcolsep}{4pt}
    \scalebox{0.9}{\begin{tabular}{l|c|ccc|c}
    $\mathcal{F}_{iam}$ & act. & \aps & \textit{t} (ms) \\
    \tline
    $3\!\times\!3$ conv & sigmoid & 32.0 & 51.9 & 33.5 & 22.9 \\
    $3\!\times\!3$ conv & softmax & 31.6 & 51.4 & 32.9 & 22.9 \\
    $1\!\times\!1$ conv & sigmoid & 30.8 & 50.7 & 32.0 & \tb{22.4} \\
    \hline
    $3\!\times\!3$ conv, ReLU, $3\!\times\!3$ conv & sigmoid & 31.9 & 52.2 & 33.0 & 23.6 \\
    Group $3\!\times\!3$ conv (2 groups) & sigmoid & 32.2 & 52.3 & 33.5 & 23.1 \\
    Group $3\!\times\!3$ conv (4 groups) & sigmoid & \tb{32.7} & \tb{53.1} & \tb{34.0} & 23.3 \\
    \end{tabular}}
    \vspace{-5pt}
    \caption{\textbf{Ablation on $\mathcal{F}_{iam}$.} Using softmax or $1\!\times\!1$ conv brings 0.4 AP and 1.2 AP drop respectively, and using two $3\!\times\!3$ conv with ReLU brings no gain. However, Group-IAM with 4 groups obtains 0.7 AP improvement.}
    \label{tab:ablation_iam}
    \vspace{-10pt}
\end{table}

\begin{table*}
\centering
\begin{minipage}[t]{0.333\linewidth}
    \centering
    \renewcommand\arraystretch{1.1}
    \renewcommand\tabcolsep{3.5pt}
    \small
    \scalebox{0.9}{
    \begin{tabular}{c|cc|cccc}
    Dice & Focal & BCE  & \aps & AP$_L$\\
    \tline
     & & \checkmark  & 23.9 & 40.2 & 24.3 & 40.8 \\
    \checkmark &  &  & 31.0 & 50.8 & 32.0 & 46.4\\
    \checkmark & \checkmark & & 31.5 & 51.6 & 32.7 & 47.5\\
    \checkmark & & \checkmark & \tb{32.0} & \tb{52.0} & \tb{33.3} & \tb{48.2} \\
    \end{tabular}}
    \caption{\textbf{Ablation on the hybrid mask loss.} We evaluate the effects of the different hybrid mask loss. Dice loss is an essential component and adding extra BCE loss can further improve the performance (+1.0 AP) especially for larger objects (+1.8 AP$_L$).}
    \label{tab:ablation_mask_loss}
\end{minipage}
\hfill
\begin{minipage}[t]{0.313\linewidth}
    \centering
    \renewcommand\arraystretch{1.1}
    \renewcommand\tabcolsep{3.5pt}
    \small
    \scalebox{0.9}{
    \begin{tabular}{c|c|c|ccc}
    w/ obj. & rescore? & loss & \aps\\
    \tline
    \xmark & - & - & 30.7 & 51.3 & 31.6 \\
    \checkmark & \xmark & CE & 31.4 & \tb{52.1} & 32.2 \\
    \checkmark & \checkmark & CE & \tb{32.0} & 52.0 & \tb{33.3} \\
    \checkmark & \checkmark & L1 & 31.5 & 51.3 & 32.7  \\
    \end{tabular}}
    \caption{\textbf{Ablation on the IoU-aware objectness.} Adding objectness facilitates more instance-aware features and improves the performance even without rescoring. Using cross-entropy loss obtains better results than L1 loss.}
    \label{tab:ablation_iou_objectness}
\end{minipage}
\hfill
\begin{minipage}[t]{0.310\linewidth}
    \centering
    \renewcommand\arraystretch{1.1}
    \renewcommand\tabcolsep{3pt}
    \small
    \scalebox{0.90}{
    \begin{tabular}{l|ccc|c}
    $\mathcal{F}_{iam}$ & \aps & \textit{t}~(ms)\\
    \tline
    $1\!\times\!1$ conv & 30.8 & 50.7 & 32.0 & \tb{22.4} \\
    $3\!\times\!3$ conv & 32.0 & 51.9 & 33.5 & 22.9 \\
    Group $3\!\times\!3$ conv & \tb{32.7} & \tb{53.1} & \tb{34.0} & 23.3 \\
    \hline
    Cross Attention & 31.8 & 51.7 & 33.1 & 23.4 \\
    \end{tabular}}
    \caption{\textbf{Comparison with cross attention.} 
    We evaluate the performance of directly using one 4-head cross attention~\cite{DETRCarionMSUKZ20} with 100 queries to segment objects. Notably, (Group-) IAM with $3\!\times\!3$ conv can offer better results}
    \label{tab:ablation_cross_attention}
\end{minipage}
\vspace{-8pt}
\end{table*}

\paragraph{Hybrid Mask Loss.}
In Table~\ref{tab:ablation_mask_loss}, we analyze the effects of the hybrid mask loss.
Notably, dice loss is the critical component for mask prediction and removing dice loss lead to the collapse (AP rapidly drops 8.1 points).
Compared to RoI-based methods~\cite{MaskRCNNHeGDG17}, full-resolution instance segmentation has severe imbalance problem between background and foreground, especially for small objects which may occupy less than 0.5\% pixels.
Dice loss is more robust to the foreground/background imbalance thus effective to handle the full-resolution segmentation.
In Table~\ref{tab:ablation_mask_loss}, adding a pixel-wise classification loss can further improve the segmentation accuracy: using binary cross-entropy loss (BCE) or focal loss improves by 1.0 AP and 0.5 AP respectively. 
Moreover, we note that pixel-wise loss significantly improves AP$_L$ (\eg, +1.8 AP from BCE) for large objects.
Additionally, increasing the weight for pixel-wise loss ($\lambda_{pix}$), \eg, 5.0, will bring some improvements.
\paragraph{IoU-aware Objectness.} 
We further conduct ablations to investigate the effects of the proposed IoU-aware objectness method. 
In Table~\ref{tab:ablation_iou_objectness}, employing the IoU-aware objectness can improve the baseline by 1.3 AP.
Interestingly, we observe that adding objectness prediction without rescoring still brings 0.7 AP improvements, which has no direct impact to classification or segmentation.
The targets for objectness differs among foreground instances and therefore the objectness loss can facilitate the instance branch to learn more instance-aware features for distinguishing objects as discussed in $\S$\ref{iou_objectness}.
We also compare different types of loss, \ie, L1 loss and cross-entropy, for IoU-aware objectness and Table~\ref{tab:ablation_iou_objectness} shows the superiority of using cross-entropy.


\subsection{Timing}
\begin{table}
    \centering
    \renewcommand\arraystretch{1.1}
    \small
    \setlength{\tabcolsep}{4pt}
    \scalebox{0.9}{\begin{tabular}{c|cccc}
    size & backbone & encoder & decoder & post \\
    \tline
    512 & 10.0 (54.3\%) & 2.5 (13.5\%) & 4.1 (22.2\%) & 1.8 (10.0\%) \\
    640 & 13.3 (55.6\%) & 2.9 (12.1\%) & 5.6 (23.4\%) & 2.1 (8.90\%)  \\
    \end{tabular}}
    \caption{\textbf{Inference time.} We report the inference latency of module of the SparseInst. The backbone consumes more than 50\% of the total time.}
    \label{tab:inference_time}
    \vspace{-12pt}
\end{table}

Our framework achieves fast inference speed for since it saves much computation costs by using single-level prediction, highlighting a sparse set of instances, fully convolutional design, and adopting extremely simple post-processing without sorting or NMS.
To better understand the efficiency of the proposed method, we measure the inference latency of each module (\ie, backbone, encoder, decoder, and post-processing). We disable the asynchronous execution in GPU for accurately recording the time, which slows down the overall inference speed.
Table~\ref{tab:inference_time} shows the inference latency (ms) of each module in SparseInst with different input resolutions.
It's worth noting that the backbone (\ie, ResNet-50) consumes most of the inference time and the post-processing inevitably requires nearly 2ms to process the final segmentation and recognition results for evaluation. 
The $3\times3$ convolutions in the decoder take much time and can be pruned for more efficient inference.

\subsection{Comparison with Cross Attention}
The proposed IAM has some connections with query-based methods~\cite{DETRCarionMSUKZ20,WangZAYC21MaxDeepLab,mask_former_abs-2107-06278,KNetabs-2106-14855}.
The cross attention between object queries $\mathbf{Q}$ and image features $\mathbf{X}$ can be briefly formulated by: $\mathbf{A}\!=\! \mathbf{Q}\mathbf{X}$ and $\mathbf{O}\!= \!\text{Softmax}(\mathbf{A})\mathbf{X}^T$, where $\mathbf{A}$ and $\mathbf{O}$ are attention maps and output queries. The cross attention has similar formulations with IAM in $\S$\ref{sec:instance_activaion_maps} especially for $1\!\times\!1$ conv, which can be viewed as 1-head cross attention.
Differently, we adopt the $3\!\times\!3$ conv as $\mathcal{F}_{iam}$ to highlight object regions, which acts as a direct spatial object representation.
Compared to queries or $1\!\times\!1$ conv, $3\!\times\!3$ conv perceives larger context and local patterns for instance recognition.
Further, we replace IAM with a 4-head cross attention and 100 queries to generate instance features, and Table~\ref{tab:ablation_cross_attention} shows that the 4-head cross attention drops 0.2 AP or 0.9 AP compared to IAM and Group-IAM, respectively.

\subsection{Visualizations}
\paragraph{Instance Activation Maps.}
\begin{figure*}[h]
    \centering
    \includegraphics[width=0.95\linewidth]{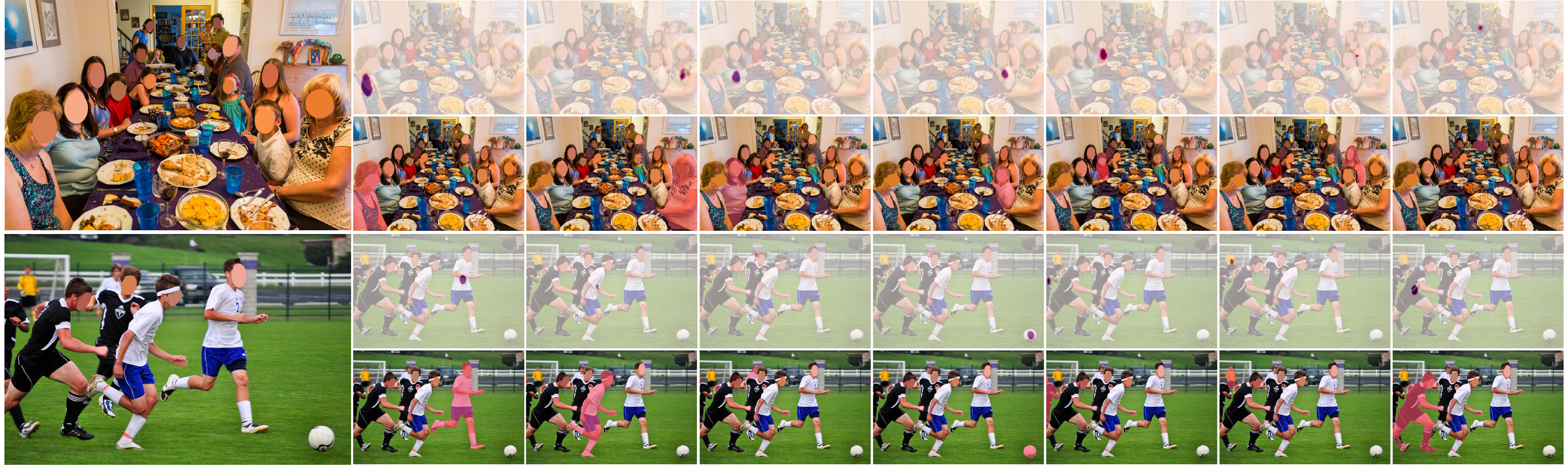}
    \vspace{-5pt}
    \caption{\textbf{Visualizations for Instance Activation Maps.} We present the visualizations of the instance activation maps and segmentation masks. For each input image, the upper row shows the instance activation maps and the bottom row shows the corresponding segmentation masks. The instance activation maps tend to highlight the discriminative regions of the objects regardless of the scales, occlusion, and poses.
    Best viewed on screen after zooming in.}
    \label{fig:vis_iam_mask}
    \vspace{-5pt}
\end{figure*}
Figure~\ref{fig:vis_iam_mask} provides the visualizations for instance activation maps and corresponding segmentation masks. Each instance activation map highlights a prominent region of the object.
Segmentation masks are well-localized and aligned with the instance activation maps.
Moreover, instance activation maps can highlight objects in despite of the scales, positions, categories and also perform well for crowd scenes.

\begin{figure*}[htbp]
    \centering
    \includegraphics[width=0.95\linewidth]{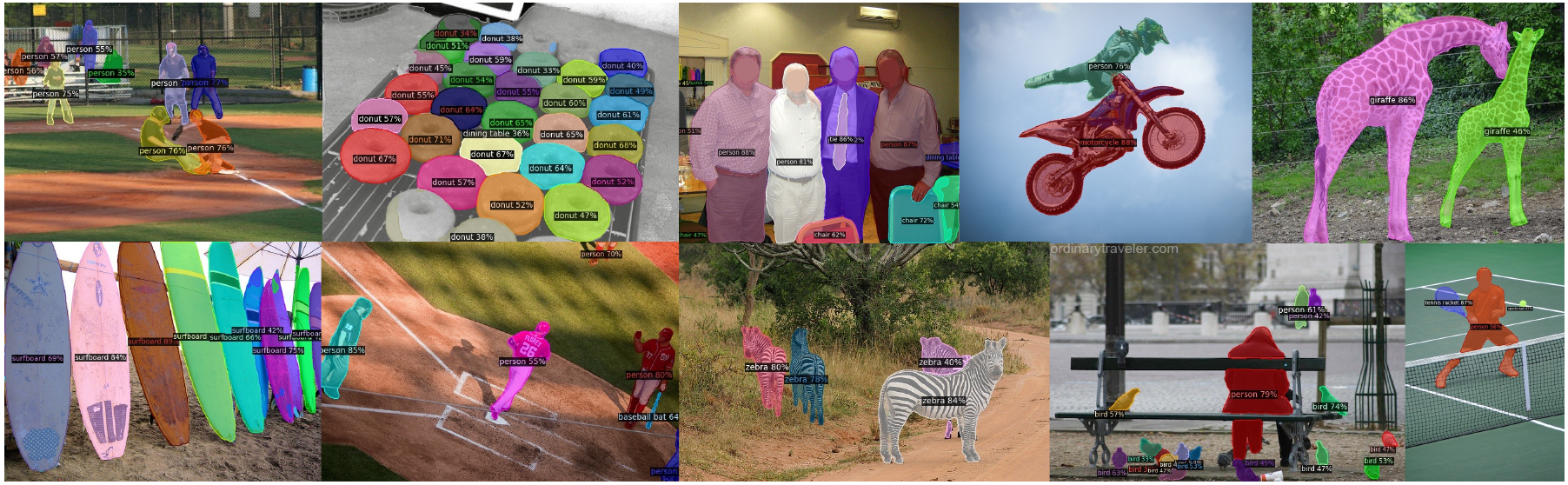}
    \vspace{-5pt}
    \caption{\textbf{Visualizations for Instance Segmentation.} The results are obtained by \name{} on COCO \texttt{val2017}. The confidence threshold is set to 0.4. We can observe that \name~can generate precise boundaries, highlight and segment well on the crowd scenes, and cope with the scale-variant segmentation.}
    \vspace{-5pt}
    \label{fig:vis_mask}
\end{figure*}

\begin{figure}[]
\centering
\includegraphics[width=0.95\linewidth]{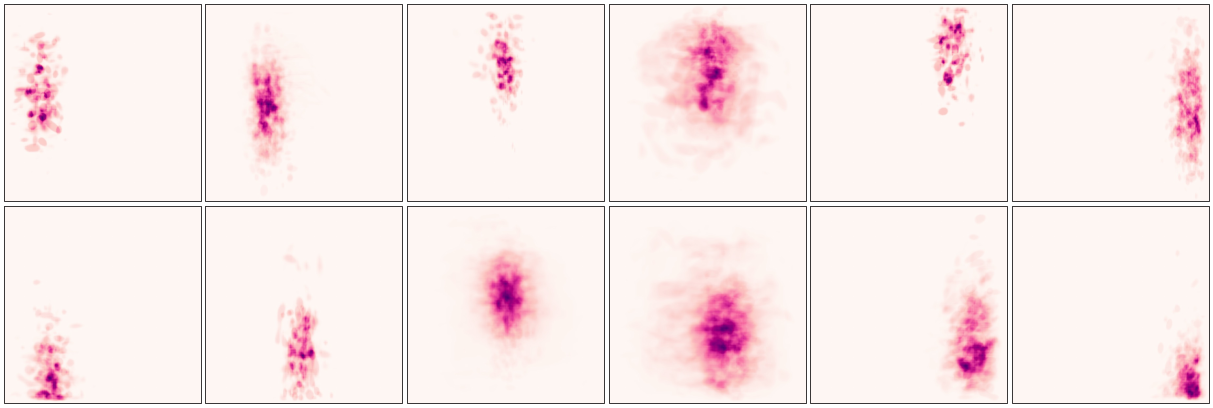}
\caption{\textbf{Visualizations for Instance Activation Maps over the COCO dataset.} We gather the 100 instance activation maps over the 5,000 images from the COCO \texttt{val2017} by averaging the activation responses for each map. Instance activation maps from different images are resized to the same size $512\times512$. We provide 12 instance activation maps for visualization.}
\label{fig:vis_iam_all}
\vspace{-10pt}
\end{figure}

For a better understanding of how the instance activation maps can discriminate objects, we further provide the visualizations of the instance activation maps from all images. Figure~\ref{fig:vis_iam_all} illustrates 12 (of 100) instance activation maps by averaging the activation response over the 5,000 images from COCO \texttt{val2017}. Different instance activation maps highlight regions of different spatial locations, scales, and shapes, which contributes to separating the instances of the same or different categories.
\paragraph{Qualitative Results.}
Figure~\ref{fig:vis_mask} shows the qualitative results of \name. The proposed \name~can generate precise segmentation masks with fine boundaries. For crowd and dense scenes, \name~can also distinguish different instances well.

\section{Conclusion}
In this work, we have explored a novel object representation by instance activation maps, which are instance-aware weighted maps and aim to highlight informative regions of objects.
Then we present a new \textit{highlight to segment} paradigm to exploit a sparse set of instance activation maps to highlight objects and aggregate instance features according to the activation maps for instance-level recognition and segmentation.
Following this paradigm, we propose \name{}, a conceptually novel and efficient end-to-end framework, which  achieves rather fast inference speed with highly competitive accuracy for real-time instance segmentation.
Extensive experiments and qualitative results have demonstrated the effectiveness of the core idea as well as the superiority of the trade-off between speed and accuracy.
Finally, we hope that \name{} can serve as a general framework for end-to-end real-time instance segmentation and be applied to practical scenes for its effectiveness and efficiency.
\paragraph{Acknowledgement.} This work was in part supported by NSFC (No. 61876212 and No. 61733007) and CAAI-Huawei MindSpore Open Fund.





{\small
\bibliographystyle{ieee_fullname}
\bibliography{egbib}
}

\newpage
\section*{Appendix}
\section*{A.1. TIDE Error Analysis}
\begin{figure}[h]
    \centering
    \includegraphics[width=0.9\linewidth]{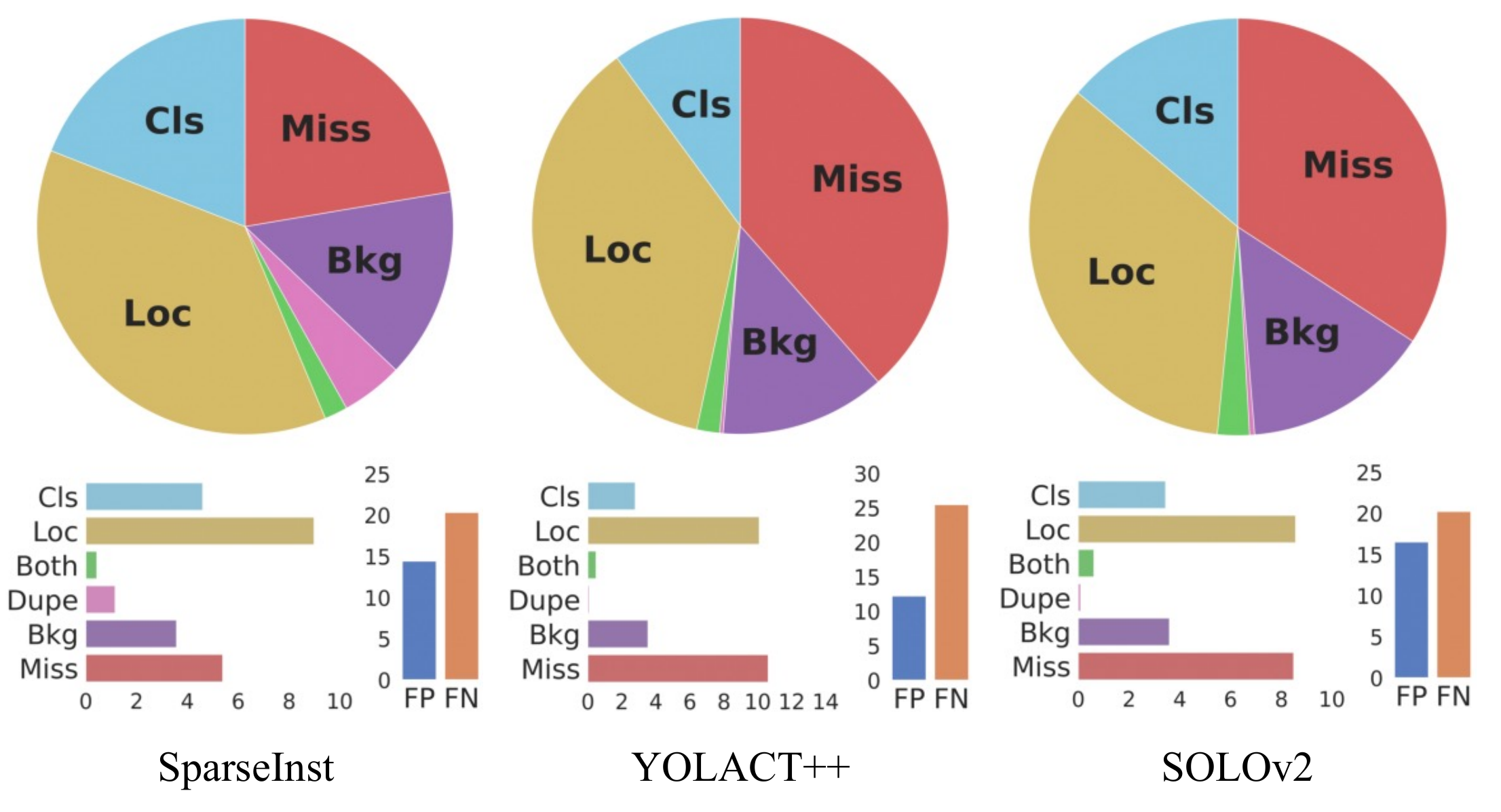}
    \caption{\textbf{TIDE Error Analysis.} We adopt TIDE~\cite{tide-eccv2020} to further analyze the errors of SparseInst, YOLACT++, and SOLOv2. \textit{cls}: classification error; \textit{loc}: localization error; \textit{miss}: missing detections; \textit{bkg}: background detections; \textit{dupe}: duplicated detections; \textit{both}: \textit{cls}+\textit{loc} error.}
    \label{fig:tide}
\end{figure}

Figure~\ref{fig:tide} shows the error analysis through TIDE~\cite{tide-eccv2020} and comparisons among SparseInst without Group-IAM (40.2FPS, 36.9AP), YOLACT++\cite{YolactBolyaZXL19} (38.6FPS, 34.1AP), and SOLOv2\cite{SOLOV2WangZKLS20} (38.2FPS, 34.0AP).
In detail, the proposed SparseInst has lower \textit{miss} error, indicating that SparseInst can discover more objects.
We observe that SparseInst has higher portions of \textit{classification} error and \textit{dupe} error than YOLACT++ or SOLOv2, and the two types of error can be attributed to classification.
SparseInst removes duplicate predictions through classification scores and better classification capability can offer better performance.

\end{document}